\documentclass[12pt]{article}
\def\,{\mskip 3mu} \def\>{\mskip 4mu plus 2mu minus 4mu} \def\;{\mskip 5mu plus 5mu} \def\!{\mskip-3mu}
\def\dispmuskip{\thinmuskip= 3mu plus 0mu minus 2mu \medmuskip=  4mu plus 2mu minus 2mu \thickmuskip=5mu plus 5mu minus 2mu}
\def\textmuskip{\thinmuskip= 0mu                    \medmuskip=  1mu plus 1mu minus 1mu \thickmuskip=2mu plus 3mu minus 1mu}
\textmuskip
\def\beq{\dispmuskip\begin{equation}}    \def\eeq{\end{equation}\textmuskip}
\def\beqn{\dispmuskip\begin{displaymath}}\def\eeqn{\end{displaymath}\textmuskip}
\def\bqa{\dispmuskip\begin{eqnarray}}    \def\eqa{\end{eqnarray}\textmuskip}
\def\bqan{\dispmuskip\begin{eqnarray*}}  \def\eqan{\end{eqnarray*}\textmuskip}

\newenvironment{keywords}{\centerline{\small\bf
Keywords}\vspace{0.5ex}\begin{quote}\small}{\par\end{quote}\vskip
1ex}
\def\nq{\hspace{-1em}}
\def\odt{{\textstyle{1\over 2}}}

\def\vec#1{{\bf #1}}
\def\p{{\scriptscriptstyle+}}
\def\pp{{\scriptscriptstyle++}}
\def\n{n}
\def\npp{\n}
\def\t{\pi}

\begin{document}

%%%%%%%%%%%%%%%%%%%%%%%%%%%%%%%%%%%%%%%%%%%%%%%%%%%%%%%%%%%%%%%%%
%                      T i t l e - P a g e                      %
%%%%%%%%%%%%%%%%%%%%%%%%%%%%%%%%%%%%%%%%%%%%%%%%%%%%%%%%%%%%%%%%%

\begin{titlepage}

\begin{center}
 {\small Technical Report IDSIA-13-01 \hfill 15 December 2001}\\[5mm]
  {\Large\sc\hrule height1pt \vskip 2mm
     Distribution of Mutual Information
     \vskip 5mm \hrule height1pt} \vspace{10mm}
  {\bf Marcus Hutter} \\[10mm]
  {\rm IDSIA, Galleria 2, CH-6928 Manno-Lugano, Switzerland}  \\
  {\rm\footnotesize marcus@idsia.ch \qquad
      http://www.idsia.ch/$^{_{_\sim}}\!$marcus}
      \\[15mm]
\end{center}


\begin{abstract}
The mutual information of two random variables $\imath$ and
$\jmath$ with joint probabilities $\{\t_{ij}\}$ is commonly used in
learning Bayesian nets as well as in many other fields. The
chances $\t_{ij}$ are usually estimated by the empirical sampling
frequency $\n_{ij}/\n$ leading to a point estimate $I(\n_{ij}/\n)$
for the mutual information. To answer questions like ``is
$I(\n_{ij}/\n)$ consistent with zero?'' or ``what is the
probability that the true mutual information is much larger than
the point estimate?'' one has to go beyond the point estimate.
In the Bayesian framework one can answer
these questions by utilizing a (second order) prior distribution
$p(\t)$ comprising prior information about $\t$. From the prior
$p(\t)$ one can compute the posterior $p(\t|\vec\n)$, from which
the distribution $p(I|\vec\n)$ of the mutual information can be
calculated.
We derive reliable and quickly computable approximations for
$p(I|\vec\n)$. We concentrate on the mean, variance, skewness, and
kurtosis, and non-informative priors. For the mean we also give an
exact expression. Numerical issues and the range of validity are
discussed.
\end{abstract}

\end{titlepage}

%%%%%%%%%%%%%%%%%%%%%%%%%%%%%%%%%%%%%%%%%%%%%%%%%%%%%%%%%%%%%%%
\section{Introduction}\label{secInt}
%%%%%%%%%%%%%%%%%%%%%%%%%%%%%%%%%%%%%%%%%%%%%%%%%%%%%%%%%%%%%%%
The mutual information $I$ (also called cross entropy) is a
widely used information theoretic measure for the stochastic
dependency of random variables \cite{Cover:91,Soofi:00}. It is
used, for instance, in learning Bayesian nets
\cite{Buntine:96,Heckerman:98}, where stochastically dependent
nodes shall be connected. The mutual information defined in
(\ref{mi}) can be computed if the joint probabilities
$\{\t_{ij}\}$ of the two random variables $\imath$ and
$\jmath$ are known. The standard procedure in the common case
of unknown chances $\t_{ij}$ is to use the sample frequency
estimates ${\n_{ij}\over\n}$ instead, as if they were
precisely known probabilities; but this is not always
appropriate. Furthermore, the point estimate
$I({\n_{ij}\over\n})$ gives no clue about the reliability of
the value if the sample size $n$ is finite. For instance, for
independent $\imath$ and $\jmath$, $I(\t)=0$ but
$I({\n_{ij}\over\n})=O(n^{-1/2})$ due to noise in the data.
The criterion for judging dependency is how many standard
deviations $I({\n_{ij}\over\n})$ is away from zero. In
\cite{Kleiter:96,Kleiter:99} the probability that the true
$I(\vec\t)$ is greater than a given threshold has been used to
construct Bayesian nets. In the Bayesian framework one can
answer these questions by utilizing a (second order) prior
distribution
$p(\t)$%
%comprising prior information about $\t$.
,which takes account of any impreciseness about $\t$.
From the prior
$p(\t)$ one can compute the posterior $p(\t|\vec n)$, from which
the distribution $p(I|\vec\n)$ of the mutual information can be
obtained.

The objective of this work is to derive reliable and quickly
computable analytical expressions for $p(I|\vec\n)$. Section
\ref{secMI} introduces the mutual information distribution,
Section \ref{secResults} discusses some results in advance before
delving into the derivation. Since the central limit theorem
ensures that $p(I|\vec\n)$ converges to a Gaussian distribution a
good starting point is to compute the mean and variance of
$p(I|\vec\n)$. In section \ref{secApprox} we relate the mean and
variance to the covariance structure of $p(\t|\vec n)$. Most
non-informative priors lead to a Dirichlet posterior. An exact
expression for the mean (Section \ref{secExact}) and approximate
expressions for the variance (Sections \ref{secDD}) are given for
the Dirichlet distribution. More accurate estimates of the
variance and higher central moments are derived in Section
\ref{secGeneral}, which lead to good approximations of
$p(I|\vec\n)$ even for small sample sizes. We show that the
expressions obtained in \cite{Kleiter:96,Kleiter:99} by heuristic
numerical methods are incorrect. Numerical issues and the range of
validity are briefly discussed in section \ref{secNum}.

%%%%%%%%%%%%%%%%%%%%%%%%%%%%%%%%%%%%%%%%%%%%%%%%%%%%%%%%%%%%%%%
\section{Mutual Information Distribution}\label{secMI}
%%%%%%%%%%%%%%%%%%%%%%%%%%%%%%%%%%%%%%%%%%%%%%%%%%%%%%%%%%%%%%%

We consider discrete random variables $\imath\in\{1,...,r\}$ and $\jmath\in
\{1,...,s\}$ and an i.i.d.\ random process with samples
$(i,j)\in\{1,...,r\}\times\{1,...,s\}$ drawn with joint probability
$\t_{ij}$. An important measure of the stochastic
dependence of $\imath$ and $\jmath$ is the mutual
information
\beq\label{mi}
  I({\vec \t}) \;=\; \sum_{i=1}^r\sum_{j=1}^s
  \t_{ij}\log{\t_{ij}\over\t_{i\p}\t_{\p j}} \;=\;
  \sum_{ij}\t_{ij}\log\t_{ij} -
  \sum_{i}\t_{i\p}\log\t_{i\p} -
  \sum_{j}\t_{\p j}\log\t_{\p j}.
\eeq
$\log$ denotes the natural logarithm and
$\t_{i\p}=\sum_j\t_{ij}$ and
$\t_{\p j}=\sum_i\t_{ij}$ are marginal probabilities.
Often one does not know the probabilities $\t_{ij}$ exactly,
but one has a sample set with $\n_{ij}$ outcomes of pair $(i,j)$.
The frequency $\hat\t_{ij}:={\n_{ij}\over\npp}$ may
be used as a first estimate of the unknown probabilities.
$\npp:=\sum_{ij}\n_{ij}$ is the total sample size.
This leads to a point (frequency) estimate $I(\hat\vec\t) =
\sum_{ij}{\n_{ij}\over\npp}
\log{\n_{ij}\npp\over\n_{i\p}\n_{\p j}}$
for the mutual information (per sample).

Unfortunately the point estimation $I(\hat\vec\t)$ gives no
information about its accuracy. In the Bayesian approach to this
problem one assumes a prior (second order) probability density
$p(\vec\t)$ for the unknown probabilities $\t_{ij}$ on the
probability simplex. From this one can compute the posterior
distribution $p(\vec\t|\vec\n) \propto
p(\t)\prod_{ij}\t_{ij}^{\n_{ij}}$ (the $n_{ij}$ are multinomially
distributed). This allows to compute the
posterior probability density of the mutual information.$\!$%
\footnote{$I(\vec\t)$ denotes the mutual information for the
specific chances $\vec\t$, whereas $I$ in the context above is
just some non-negative real number. $I$ will also denote the
mutual information {\it random variable} in the
expectation $E[I]$ and variance $\mbox{Var}[I]$. Expectaions are
{\it always} w.r.t.\ to the posterior distribution
$p(\vec\t|\vec\n)$. }
\beq\label{midistr}
  p(I|\vec\n) = \int
  \delta(I(\vec\t)-I)p(\vec\t|\vec\n)d^{rs}\vec\t
\eeq
\footnote{Since $0\leq I(\t)\leq I_{max}$ with sharp upper
bound $I_{max}:= \min\{\log r,\log s\}$, the integral may be
restricted to $\int_0^{I_{max}}$, which shows that the domain
of $p(I|\vec n)$ is $[0,I_{max}].$}%
The $\delta()$ distribution restricts the integral to $\t$ for
which $I(\t)=I$. For large sample size $\npp\to\infty$,
$p(\vec\t|\vec\n)$ is strongly peaked around $\vec\t=\hat\vec\t$
and $p(I|\vec\n)$ gets strongly peaked around the frequency
estimate $I=I(\hat\vec\t)$. The mean $E[I] = \int_0^\infty I
p(I|\vec\n)\,dI = \int I(\vec\t)p(\vec\t|\vec\n)d^{rs}\vec\t$ and
the variance $\mbox{Var}[I]=E[(I-E[I])^2]=E[I^2]-E[I]^2$ are of
central interest.

%%%%%%%%%%%%%%%%%%%%%%%%%%%%%%%%%%%%%%%%%%%%%%%%%%%%%%%%%%%%%%%
\section{Results for $I$ under the Dirichlet P{\rm(}oste{\rm)}rior}\label{secResults}
%%%%%%%%%%%%%%%%%%%%%%%%%%%%%%%%%%%%%%%%%%%%%%%%%%%%%%%%%%%%%%%
Most\footnote{But not all priors which one can argue to be
non-informative lead to Dirichlet posteriors. Brand \cite{Brand:99}
(and others), for instance, advocate the entropic prior
$p(\vec\t)\propto e^{-H(\vec\t)}$.}
non-informative priors for $p(\t)$ lead to a Dirichlet
posterior distribution $p(\vec\t|\vec\n) \propto
\prod_{ij}\t_{ij}^{\n_{ij}-1}$ with interpretation
$\n_{ij}=\n'_{ij}+\n''_{ij}$, where
$\n'_{ij}$ are the number of samples $(i,j)$, and
$\n''_{ij}$ comprises prior information
($1$ for the uniform prior, $\odt$ for Jeffreys' prior, $0$ for
Haldane's prior, ${1\over rs}$ for Perks' prior \cite{Gelman:95}).
In principle this allows to compute the
posterior density $p(I|\vec\n)$ of the mutual information. In
sections \ref{secApprox} and \ref{secDD} we expand the mean and
variance in terms of $\npp^{-1}$:
\bqa\label{mvappr}
  E[I] &=&
  \sum_{ij}{\n_{ij}\over\npp}
  \log{\n_{ij}\npp\over\n_{i\p}\n_{\p j}} \;+\;
  {(r-1)(s-1)\over 2\npp} \;+\; O(\npp^{-2}),
  \\\nonumber
  \mbox{Var}[I] &=&
  {1\over\npp}
  \sum_{ij}{\n_{ij}\over\npp}\bigg(\log{\n_{ij}\npp\over
    \n_{i\p}\n_{\p j}}\bigg)^2 -
  {1\over\npp}\bigg(\sum_{ij}{\n_{ij}\over\npp}\log{\n_{ij}\npp\over
    \n_{i\p}\n_{\p j}}\bigg)^2 \;+\; O(\npp^{-2}).
\eqa
The first term for the mean is just the point estimate
$I(\hat\t)$. The second term is a small correction if $\npp\gg
r \cdot s$. Kleiter \cite{Kleiter:96,Kleiter:99} determined the
correction by Monte Carlo studies as $\min\{{r-1\over
2\npp},{s-1\over 2\npp}\}$. This is wrong unless $s$ or $r$ are 2.
The expression $2E[I]/n$ they determined for the variance has a
completely different structure than ours. Note that the mean is
lower bounded by ${const.\over\npp}+O(\npp^{-2})$, which is
strictly positive for large, but finite sample sizes, even if
$\imath$ and $\jmath$ are statistically independent and
independence is perfectly represented in the data ($I(\hat\t)=0$).
On the other hand, in this case, the standard deviation
$\sigma=\sqrt{\mbox{Var} (I)}\sim {1\over\npp}\sim E[I]$ correctly
indicates that the mean is still consistent with zero.

Our approximations (\ref{mvappr}) for the mean and variance are
good if ${r \cdot s\over\npp}$ is small. The central limit
theorem ensures that $p(I|\vec\n)$ converges to a Gaussian
distribution with mean $E[I]$ and variance $\mbox{Var}[I]$. Since
$I$ is non-negative it is more appropriate to approximate
$p(I|\vec\t)$ as a Gamma ($=$ scaled $\chi^2$) or log-normal
distribution with mean $E[I]$ and variance $\mbox{Var}[I]$, which
is of course also asymptotically correct.

A systematic expansion in $\npp^{-1}$ of the mean, variance, and
higher moments is possible but gets arbitrarily cumbersome.
The $O(\npp^{-2})$ terms for the variance and leading order
terms for the skewness and kurtosis
are given in Section \ref{secGeneral}.
For the mean it is possible to give an exact expression
\beq\label{miexex2}
  E[I] = {1\over\npp}\sum_{ij}\n_{ij}
  [\psi(\n_{ij}+1)-\psi(\n_{i\p}+1)-\psi(\n_{\p
  j}+1)+\psi(\npp+1)]
\eeq
with $\psi(n+1)=-\gamma+\sum_{k=1}^n{1\over k}=\log
n+O({1\over n})$ for integer $n$. See Section \ref{secExact} for
details and more general expressions for $\psi$ for non-integer
arguments.

There may be other prior information available which cannot be
comprised in a Dirichlet distribution. In this general case, the
mean and variance of $I$ can still be related to the covariance
structure of $p(\t|\vec\n)$, which will be done in the following
Section.

%%%%%%%%%%%%%%%%%%%%%%%%%%%%%%%%%%%%%%%%%%%%%%%%%%%%%%%%%%%%%%%
\section{Approximation of Expectation and Variance of $I$}\label{secApprox}
%%%%%%%%%%%%%%%%%%%%%%%%%%%%%%%%%%%%%%%%%%%%%%%%%%%%%%%%%%%%%%%
In the following let $\hat\t_{ij}:=E[\t_{ij}]$.
Since $p(\vec\t|\vec\n)$ is strongly peaked
around $\vec\t=\hat\vec\t$ for large $\npp$ we may
expand $I(\t)$ around $\hat\vec\t$ in the integrals for the mean and the variance.
With
$\Delta_{ij}:=\t_{ij}-\hat\t_{ij}$ and using $\sum_{ij}\t_{ij}= 1
=\sum_{ij}\hat\t_{ij}$ we get for the expansion of (\ref{mi})
\beq\label{miexp}
  I(\t) \;=\; I(\hat\t) +
  \sum_{ij}\log\left({\hat\t_{ij}\over\hat\t_{i\p}\hat\t_{\p j}}\right)\Delta_{ij}
  + \sum_{ij}{\Delta_{ij}^2\over 2\hat\t_{ij}} -
  \sum_i{\Delta_{i\p}^2\over 2\hat\t_{i\p}} -
  \sum_j{\Delta_{\p j}^2\over 2\hat\t_{\p j}} +
  O(\Delta^3).
\eeq
Taking the expectation, the linear term $E[\Delta_{ij}]=0$ drops
out. The quadratic terms $E[\Delta_{ij}\Delta_{kl}] =
\mbox{Cov}(\t_{ij},\t_{kl})$ are the covariance of $\t$ under
distribution $p(\vec\t|\vec\n)$ and are proportional to
$\npp^{-1}$. It can be shown that $E[\Delta^3]\sim\npp^{-2}$ (see
Section \ref{secGeneral}).
\beq\label{exnlo}
  E[I] \;=\; I(\hat\t) + {1\over 2}
  \sum_{ijkl}\left({\delta_{ik}\delta_{jl}\over\hat\t_{ij}} -
  {\delta_{ik}\over\hat\t_{i\p}} -
  {\delta_{jl}\over\hat\t_{\p j}}\right)\mbox{Cov}(\t_{ij},\t_{kl}) +
  O(\npp^{-2}).
\eeq
The Kronecker delta $\delta_{ij}$ is $1$ for $i=j$ and $0$ otherwise.
The variance of $I$ in leading order in $\npp^{-1}$ is
\bqa\nonumber
  \mbox{Var}\,I(\t) &=&
  E[(I-E[I])^2] \;\stackrel+=\;
  E\left[\left(\sum_{ij}\log\left({\hat\t_{ij}\over
    \hat\t_{i\p}\hat\t_{\p j}}\right)\Delta_{ij}\right)^2\right]
  \;=\; \\\label{varlo}
  &=&
  \sum_{ijkl}\log{\hat\t_{ij}\over\hat\t_{i\p}\hat\t_{\p j}}
  \log{\hat\t_{kl}\over\hat\t_{k\p}\hat\t_{\p l}}
  \mbox{Cov}(\t_{ij},\t_{kl}),
\eqa
where $\stackrel+=$ means $=$ up to terms of order
$\npp^{-2}$. So the leading order variance and the leading and
next to leading order mean of the mutual information $I(\t)$ can be
expressed in terms of the covariance of $\t$ under the posterior distribution
$p(\t|\vec\n)$.

%%%%%%%%%%%%%%%%%%%%%%%%%%%%%%%%%%%%%%%%%%%%%%%%%%%%%%%%%%%%%%%
\section{The Second Order Dirichlet Distribution}\label{secDD}
%%%%%%%%%%%%%%%%%%%%%%%%%%%%%%%%%%%%%%%%%%%%%%%%%%%%%%%%%%%%%%%
Noninformative priors for $p(\t)$ are commonly used if no
additional prior information is available. Many non-informative
choices (uniform, Jeffreys', Haldane's, Perks', ... prior) lead to
a Dirichlet posterior distribution:
\bqa\nonumber
  p(\t|\vec\n) &=&
  {1\over
  N(\vec\n)}\prod_{ij}\t_{ij}^{\n_{ij}-1}\delta(\t_\pp-1)
  \quad\mbox{with normalization}
  \\\label{norm}
  N(\vec\n) &=&
  \int\prod_{ij}\t_{ij}^{\n_{ij}-1}\delta(\t_\pp-1)
  d^{rs}\t \;=\;
  {\prod_{ij}\Gamma(\n_{ij})\over\Gamma(\npp)},
\eqa
where $\Gamma$ is the Gamma function, and
$\n_{ij}=\n'_{ij}+\n''_{ij}$, where $\n'_{ij}$ are
the number of samples $(i,j)$, and $\n''_{ij}$ comprises prior
information
($1$ for the uniform prior, $\odt$ for Jeffreys' prior,
$0$ for Haldane's prior, ${1\over rs}$ for Perks' prior).
Mean and covariance of $p(\t|\vec\n)$ are
\beq\label{ecov}
  \hat\t_{ij} := E[\t_{ij}]=
  {\n_{ij}\over\npp}, \quad
  \mbox{Cov}(\t_{ij},\t_{kl}) =
  {1\over\npp+1}(\hat\t_{ij}\delta_{ik}\delta_{jl}-
  \hat\t_{ij}\hat\t_{kl})
\eeq
Inserting this into (\ref{exnlo}) and (\ref{varlo}) we get after some
algebra for the mean and variance of the mutual information
$I(\t)$ up to terms of order $\npp^{-2}$:
\bqa\label{exnlodi}
  E[I] &=& J \;+\; {(r-1)(s-1)\over 2(\npp+1)}
  \;+\; O(\npp^{-2}),
  \\\label{varlodi}
  \mbox{Var}[I] &=&
  {1\over\npp+1}(K-J^2) \;+\;
  O(\npp^{-2}), \quad
  \\\label{Jdef}
  J &:=& \sum_{ij}{\n_{ij}\over\npp}\log{\n_{ij}\npp\over
    \n_{i\p}\n_{\p j}} \;=\; I(\hat\t), \quad
  \\\label{Kdef}
  K &:=& \sum_{ij}{\n_{ij}\over\npp}\left(\log{\n_{ij}\npp\over
    \n_{i\p}\n_{\p j}}\right)^2.
\eqa
$J$ and $K$ (and $L$, $M$, $P$, $Q$ defined later) depend on
$\hat\t_{ij} = {\n_{ij}\over\npp}$ only, i.e.\ are $O(1)$ in
$\vec\n$. Strictly speaking we should expand
${1\over\npp+1}={1\over\npp}+O(\npp^{-2})$, i.e.\ drop the $+1$,
but the exact expression (\ref{ecov}) for the covariance suggests
to keep the $+1$. We compared both versions with the exact values
(from Monte-Carlo simulations) for various parameters $\vec\t$. In
most cases the expansion in ${1\over\npp+1}$ was more accurate, so
we suggest to use this variant.

%%%%%%%%%%%%%%%%%%%%%%%%%%%%%%%%%%%%%%%%%%%%%%%%%%%%%%%%%%%%%%%
\section{Exact Value for $E[I]$}\label{secExact}
%%%%%%%%%%%%%%%%%%%%%%%%%%%%%%%%%%%%%%%%%%%%%%%%%%%%%%%%%%%%%%%
It is possible to get an exact expression for the mean mutual
information $E[I]$ under the Dirichlet distribution.
By noting that $x\log x = {d\over d\beta}x^\beta|_{\beta=1}$,
($x = \{\t_{ij},\t_{i\p},\t_{\p j}\}$), one
can replace the logarithms in the last expression of
(\ref{mi}) by powers. From (\ref{norm}) we see that
$E[(\t_{ij})^\beta]={\Gamma(\n_{ij}+\beta)\Gamma(\npp)\over
\Gamma(\n_{ij})\Gamma(\npp+\beta)}$. Taking the
derivative and setting $\beta=1$ we get
\beqn
  E[\t_{ij}\log\t_{ij}] = {d\over d\beta}E[(\t_{ij})^\beta]_{\beta=1}
  = {1\over\npp}\sum_{ij}\n_{ij}[\psi(\n_{ij}+1)-\psi(\npp+1)].
\eeqn
The $\psi$ function has the following properties (see
\cite{Abramowitz:74} for details)
\beqn
  \psi(z)={d\log\Gamma(z)\over dz}={\Gamma'(z)\over\Gamma(z)},\quad
  \psi(z+1)=\log z + {1\over 2z} - {1\over 12z^2} + O({1\over z^4}),
\eeqn
\beq\label{psi2}
  \psi(n)=-\gamma+\sum_{k=1}^{n-1}{1\over k},\quad
  \psi(n+\odt)=-\gamma+2\log 2+2\sum_{k=1}^n{1\over 2k-1}.
\eeq
The value of the Euler constant $\gamma$ is irrelevant here,
since it cancels out. Since the marginal distributions of
$\t_{i\p}$ and $\t_{\p j}$ are also Dirichlet (with parameters
$\n_{i\p}$ and $\n_{\p j}$) we get similarly
\bqan
  E[\t_{i\p}\log\t_{i\p}] &=&
  {1\over\npp}\sum_i\n_{i\p}[\psi(\n_{i\p}+1)-\psi(\npp+1)],
  \\
  E[\t_{\p j}\log\t_{\p j}] &=&
  {1\over\npp}\sum_j\n_{\p j}[\psi(\n_{\p j}+1)-\psi(\npp+1)].
\eqan
Inserting this into (\ref{mi}) and rearranging terms we get the
exact expression\footnote{This expression has independently
been derived in \cite{Wolpert:93b}.}
\beq\label{miexex}
  E[I] = {1\over\npp}\sum_{ij}\n_{ij}
  [\psi(\n_{ij}+1)-\psi(\n_{i\p}+1)-\psi(\n_{\p
  j}+1)+\psi(\npp+1)]
\eeq
For large sample sizes, $\psi(z+1)\approx\log z$ and (\ref{miexex})
approaches the frequency estimate $I(\hat\t)$ as it should be.
Inserting the expansion $\psi(z+1)=\log z+{1\over 2z}+...$ into
(\ref{miexex}) we also get the correction term ${(r-1)(s-1)\over
2\npp}$ of (\ref{mvappr}).

The presented method (with some refinements) may also be used to
determine an exact expression for the variance of $I(\t)$. All but
one term can be expressed in terms of Gamma functions. The final
result after differentiating w.r.t.\ $\beta_1$ and $\beta_2$ can
be represented in terms of $\psi$ and its derivative $\psi'$. The
mixed term $E[(\t_{i\p})^{\beta_1}(\t_{\p j})^{\beta_2}]$ is more
complicated and involves confluent hypergeometric functions, which
limits its practical use \cite{Wolpert:93b}.

%%%%%%%%%%%%%%%%%%%%%%%%%%%%%%%%%%%%%%%%%%%%%%%%%%%%%%%%%%%%%%%
\section{Generalizations}\label{secGeneral}
%%%%%%%%%%%%%%%%%%%%%%%%%%%%%%%%%%%%%%%%%%%%%%%%%%%%%%%%%%%%%%%
A systematic expansion of all moments of $p(I|\vec\n)$ to arbitrary order in
$\npp^{-1}$ is possible, but gets soon quite cumbersome.
For the mean we already gave an exact expression (\ref{miexex}), so we
concentrate here on the variance, skewness and the kurtosis of $p(I|\vec\n)$.
The $3^{rd}$ and $4^{th}$
central moments of $\t$ under
the Dirichlet distribution are
\beq\label{mom3}
  E[\Delta_a\Delta_b\Delta_c] \;=\; {2\over(\npp+1)(\npp+2)}
  [2\hat\t_a\hat\t_b\hat\t_c
   - \hat\t_a\hat\t_b\delta_{bc}
   - \hat\t_b\hat\t_c\delta_{ca}
   - \hat\t_c\hat\t_a\delta_{ab}
   + \hat\t_a\delta_{ab}\delta_{bc}]
\eeq
\bqa
   E[\Delta_a\Delta_b\Delta_c\Delta_d] &=& {1\over\npp^2}
   [3\hat\t_a\hat\t_b\hat\t_c\hat\t_d
   - \hat\t_c\hat\t_d\hat\t_a\delta_{ab}
   - \hat\t_b\hat\t_d\hat\t_a\delta_{ac}
   - \hat\t_b\hat\t_c\hat\t_a\delta_{ad} \nq\\[-2ex]\nonumber
   && \qquad\qquad\qquad\; - \hat\t_a\hat\t_d\hat\t_b\delta_{bc}
   - \hat\t_a\hat\t_c\hat\t_b\delta_{bd}
   - \hat\t_a\hat\t_b\hat\t_c\delta_{cd} \nq\\\nonumber
   && \qquad\qquad\qquad\;
   + \hat\t_a\hat\t_c\delta_{ab}\delta_{cd}
   + \hat\t_a\hat\t_b\delta_{ac}\delta_{bd}
   + \hat\t_a\hat\t_b\delta_{ad}\delta_{bc}]
   +O(\npp^{-3})\nq
\eqa
with $a = ij$, $b = kl,...\in\{1,...,r\}\times\{1,...,s\}$
being double indices,
$\delta_{ab} = \delta_{ik}\delta_{jl},...$
$\hat\t_{ij}={\n_{ij}\over\npp}$.
Expanding $\Delta^k = (\t-\hat\t)^k$ in $E[\Delta_a\Delta_b...]$ leads to
expressions containing $E[\t_a\t_b...]$, which can be
computed by a case analysis of all combinations of equal/unequal
indices $a,b,c,...$ using (\ref{norm}).
Many terms cancel leading to the above expressions.
They allow to compute the order $\npp^{-2}$ term of
the variance of $I(\t)$. Again, inspection of (\ref{mom3})
suggests to expand in $[(\npp+1)(\npp+2)]^{-1}$, rather than in
$\npp^{-2}$. The variance in leading and next to leading order
is
\bqa\label{var2ndo}
  \mbox{Var}[I] %&=&
  &=& {K-J^2\over\npp+1} +
  {M+(r - 1)(s - 1)(\odt - J)-Q
  \over(\npp+1)(\npp+2)} + O(\npp^{-3})
  \\\label{Mdef}
  M &:=& \sum_{ij}
  \left({1\over\n_{ij}}-{1\over\n_{i\p}}-{1\over\n_{\p
  j}}+{1\over\npp}\right)
  \n_{ij}\log{\n_{ij}\npp\over\n_{i\p}\n_{\p j}},
  \\\label{Qdef}
  Q &:=& 1-\sum_{ij}{\n_{ij}^2\over\n_{i\p}\n_{\p j}}.
\eqa
$J$ and $K$ are defined in (\ref{Jdef}) and (\ref{Kdef}).
Note that the first term ${K-J^2\over\n+1}$ also contains second
order terms when expanded in $\npp^{-1}$. The leading order
terms for the $3^{rd}$ and $4^{th}$ central moments of $p(I|\vec\n)$ are
\bqan
  E[(I-E[I])^3] & = &
  {2\over\npp^2}[2J^3 - 3KJ + L] +
  {3\over\npp^2}[K + J^2 - P] +
  O(\npp^{-3}),
  \\
  L & := & \sum_{ij}{\n_{ij}\over\npp}\left(\log{\n_{ij}\npp\over
    \n_{i\p}\n_{\p j}}\right)^3,\quad
  P \;:=\; \sum_i{\n J_{i\p}^2\over\n_{i\p}} + \sum_j{\n J_{\p j}^2\over\n_{\p j}},
  \\
  J_{i\p} & :=&  \sum_{j}{\n_{ij}\over\npp}\log{\n_{ij}\npp\over\n_{i\p}\n_{\p
  j}}\qquad,\quad
  J_{\p j} \;:=\; \sum_{i}{\n_{ij}\over\npp}\log{\n_{ij}\npp\over\n_{i\p}\n_{\p j}},
  \\
  E[(I-E[I])^4] & = &
  {3\over\npp^2}[K-J^2]^2 + O(\npp^{-3}),
\eqan
from which the skewness and kurtosis can be obtained by dividing
by $\mbox{Var}[I]^{3/2}$ and $\mbox{Var}[I]^2$
respectively. One can see that the skewness is of order
$\npp^{-1/2}$ and the kurtosis is $3+O(\npp^{-1})$.
Significant deviation of the skewness from $0$ or the kurtosis from
$3$ would indicate a non-Gaussian $I$. They can be used to get an improved
approximation for $p(I|\vec\n)$ by making, for instance, an ansatz
\beqn
  p(I|\vec\n)\propto (1+\tilde b I+\tilde c I^2) \cdot p_0(I|\tilde\mu,\tilde\sigma^2)
\eeqn
and fitting the parameters $\tilde b$, $\tilde c$, $\tilde\mu$,
and $\tilde\sigma^2$ to the mean, variance, skewness, and kurtosis
expressions above. $p_0$ is the Normal or Gamma distribution (or
any other distribution with Gaussian limit). From this, quantiles
$p(I > I_*|\vec\n):=\int_{I_*}^\infty p(I|\vec\n)\, dI$, needed in
\cite{Kleiter:96,Kleiter:99}, can be computed. A systematic
expansion of arbitrarily high moments to arbitrarily high order in
$\npp^{-1}$ leads, in principle, to arbitrarily accurate
estimates.

%%%%%%%%%%%%%%%%%%%%%%%%%%%%%%%%%%%%%%%%%%%%%%%%%%%%%%%%%%%%%%%
\section{Numerics}\label{secNum}
%%%%%%%%%%%%%%%%%%%%%%%%%%%%%%%%%%%%%%%%%%%%%%%%%%%%%%%%%%%%%%%
%-------------------------------%
%\subsection{Implementation of $\psi(z)$}
%-------------------------------%
There are short and fast implementations of
$\psi$. The code of the Gamma function in \cite{Press:92}, for
instance, can be modified to compute the $\psi$ function. For
integer and half-integer values one may create a lookup table from
(\ref{psi2}).
%-------------------------------%
%\subsection{Computation time of (central moments)}
%-------------------------------%
The needed quantities $J$, $K$, $L$, $M$, and $Q$ (depending on $\vec
n$) involve a double sum, $P$ only a single sum, and the $r + s$
quantities $J_{i\p}$ and $J_{\p j}$ also only a single sum. Hence,
the computation time for the (central) moments is of the same
order $O(r \cdot s)$ as for the point estimate (\ref{mi}).
%-------------------------------%
%\subsection{Exact Monte Carlo}
%-------------------------------%
``Exact'' values have been obtained for representative choices of
$\t_{ij}$, $r$, $s$, and $\npp$ by Monte Carlo simulation.
The $\t_{ij}:=x_{ij}/x_\pp$ are Dirichlet distributed, if each
$x_{ij}$ follows a Gamma distribution. See \cite{Press:92} how to
sample from a Gamma distribution.
%-------------------------------%
%\subsection{Numerical accuracy of expansion}
%-------------------------------%
The variance has been expanded in ${r \cdot s\over \npp}$,
so the relative error ${\mbox{\scriptsize
Var}[I]_{approx}-\mbox{\scriptsize Var}[I]_{exact}\over
\mbox{\scriptsize Var}[I]_{exact}}$ of the approximation
(\ref{varlodi}) and (\ref{var2ndo}) are of the order of
${r \cdot s\over \npp}$ and $({r \cdot s\over \npp})^2$
respectively, {\em if} $\imath$ and $\jmath$ are dependent. If
they are independent the leading term (\ref{varlodi}) drops
itself down to order $\npp^{-2}$ resulting in a reduced
relative accuracy $O({r \cdot s\over \npp})$ of (\ref{var2ndo}).
Comparison with the Monte Carlo values confirmed an accurracy
in the range $({r \cdot s\over\npp})^{1...2}$. The mean
(\ref{miexex2}) is exact. Together with the skewness and
kurtosis we have a good description for the distribution of
the mutual information $p(I|\vec n)$ for not too small sample
bin sizes $n_{ij}$.
%-------------------------------%
%\subsection{Useful accuracy}
%-------------------------------%
We want to conclude with some notes on {\it useful} accuracy. The
hypothetical prior sample sizes $\n''_{ij}=\{0,{1\over
rs},\odt,1\}$ can all be argued to be non-informative
\cite{Gelman:95}. Since the central moments are expansions in
$\npp^{-1}$, the next to leading order term can be freely adjusted
by adjusting $\n''_{ij}\in[0...1]$.
So one may argue that anything beyond leading order is free to
will, and the leading order terms may be regarded as accurate as
we can specify our prior knowledge. On the other hand, exact
expressions have the advantage of being safe against
cancellations. For instance, leading order of $E[I]$ and $E[I^2]$
does not suffice to compute the leading order of $\mbox{Var}[I]$.

%------------------------------%
\subsubsection*{Acknowledgements}
%------------------------------%
I want to thank Ivo Kwee for valuable discussions and Marco
Zaffalon for encouraging me to investigate this topic. This work
was supported by SNF grant 2000-61847.00 to J\"urgen Schmidhuber.

%%%%%%%%%%%%%%%%%%%%%%%%%%%%%%%%%%%%%%%%%%%%%%%%%%%%%%%%%%%%%%%
%         Bibliography        %
%%%%%%%%%%%%%%%%%%%%%%%%%%%%%%%%%%%%%%%%%%%%%%%%%%%%%%%%%%%%%%%


\begin{thebibliography}{PFTV92}

\bibitem[AS74]{Abramowitz:74}
M.~Abramowitz and I.~A. Stegun, editors.
\newblock {\em Handbook of mathematical functions}.
\newblock Dover publications, inc., 1974.

\bibitem[Bra99]{Brand:99}
M.~Brand.
\newblock Structure learning in conditional probability models via an entropic
  prior and parameter extinction.
\newblock {\em Neural Computation}, 11(5):1155--1182, 1999.

\bibitem[Bun96]{Buntine:96}
W.~Buntine.
\newblock A guide to the literature on learning probabilistic networks from
  data.
\newblock {\em {IEEE} Transactions on Knowledge and Data Engineering},
  8:195--210, 1996.

\bibitem[CT91]{Cover:91}
T.~M. Cover and J.~A. Thomas.
\newblock {\em Elements of Information Theory}.
\newblock Wiley Series in Telecommunications. John Wiley \& Sons, New York, NY,
  USA, 1991.

\bibitem[GCSR95]{Gelman:95}
A.~Gelman, J.~B. Carlin, H.~S. Stern, and D.~B. Rubin.
\newblock {\em Bayesian Data Analysis.}
\newblock Chapman, 1995.

\bibitem[Hec98]{Heckerman:98}
D.~Heckerman.
\newblock A tutorial on learning with {B}ayesian networks.
\newblock {\em Learnig in Graphical Models}, pages 301--354, 1998.

\bibitem[KJ96]{Kleiter:96}
G.~D. Kleiter and R.~Jirousek.
\newblock Learning {B}ayesian networks under the control of mutual information.
\newblock {\em Proceedings of the 6th International Conference on Information
  Processing and Management of Uncertainty in Knowledge-Based Systems
  (IPMU-1996)}, pages 985--990, 1996.

\bibitem[Kle99]{Kleiter:99}
G.~D. Kleiter.
\newblock The posterior probability of {B}ayes nets with strong dependences.
\newblock {\em Soft Computing}, 3:162--173, 1999.

\bibitem[PFTV92]{Press:92}
W.~H. Press, B.~P. Flannery, S.~A. Teukolsky, and W.~T. Vetterling.
\newblock {\em Numerical Recipes in {C}: The Art of Scientific Computing}.
\newblock Cambridge University Press, Cambridge, second edition, 1992.

\bibitem[Soo00]{Soofi:00}
E.~S. Soofi.
\newblock Principal information theoretic approaches.
\newblock {\em Journal of the American Statistical Association}, 95:1349--1353,
  2000.

\bibitem[WW93]{Wolpert:93b}
D.~R. Wolf and D.~H. Wolpert.
\newblock Estimating functions of distributions from {A} finite set of samples,
  part 2: Bayes estimators for mutual information, chi-squared, covariance and
  other statistics.
\newblock Technical Report LANL-LA-UR-93-833, Los Alamos National Laboratory,
  1993.
\newblock Also Santa Fe Insitute report SFI-TR-93-07-047.

\end{thebibliography}
\end{document}